\documentclass[twocolumn,letterpaper]{IEEEtran}  

\usepackage[]{graphicx}    
\usepackage{amsmath}
\usepackage{booktabs}
\usepackage[font=bf, figurename=Figure, labelsep=period]{caption}
\usepackage[]{svg}
\usepackage[ruled,vlined]{algorithm2e}
\usepackage{subcaption}
\usepackage{float}
\usepackage{parskip}

\usepackage[noadjust]{cite}
\newcommand{\ignore}[1]{}  

\begin{document}
\title{Scheduling the NASA Deep Space Network with Deep Reinforcement Learning}

\author{%
Edwin Goh, Hamsa Shwetha Venkataram, Mark Hoffmann, Mark Johnston, Brian Wilson\\
Jet Propulsion Laboratory, California Institute of Technology\\
4800 Oak Grove Dr., Pasadena, CA 91109\\
edwin.y.goh@jpl.nasa.gov
\thanks{\footnotesize \copyright2021 IEEE. Personal use of this material is permitted. Permission from IEEE must be obtained for all other uses, in any current or future media, including reprinting/republishing this material for advertising or promotional purposes, creating new collective works, for resale or redistribution to servers or lists, or reuse of any copyrighted component of this work in other works.}              
}

\maketitle

\thispagestyle{plain}
\pagestyle{plain}

\maketitle
\thispagestyle{plain}
\pagestyle{plain}

\begin{abstract}

With three complexes spread evenly across the Earth, NASA’s Deep Space Network (DSN) is the primary means of communications as well as a significant scientific instrument for dozens of active missions around the world. A rapidly rising number of spacecraft and increasingly complex scientific instruments with higher bandwidth requirements have resulted in demand that exceeds the network's capacity across its 12 antennae. The existing DSN scheduling process operates on a rolling weekly basis and is time-consuming; for a given week, generation of the final baseline schedule of spacecraft tracking passes takes roughly 5 months from the initial requirements submission deadline, with several weeks of peer-to-peer negotiations in between. This paper proposes a deep reinforcement learning (RL) approach to generate candidate DSN schedules from mission requests and spacecraft ephemeris data with demonstrated capability to address real-world operational constraints. A deep RL agent is developed that takes mission requests for a given week as input, and interacts with a DSN scheduling environment to allocate tracks such that its reward signal is maximized. A comparison is made between an agent trained using Proximal Policy Optimization and its random, untrained counterpart. The results represent a proof-of-concept that, given a well-shaped reward signal, a deep RL agent can learn the complex heuristics used by experts to schedule the DSN. A trained agent can potentially be used to generate candidate schedules to bootstrap the scheduling process and thus reduce the turnaround cycle for DSN scheduling. 

\end{abstract}

\tableofcontents

\section{Introduction}
\label{sec:intro}
As humankind progresses towards groundbreaking space explorations ranging from searching for signs of extraterrestrial life by roving on the red planet \cite{MARS2020} to understanding the composition interstellar space \cite{Voyager}, 
communicating with the spacecraft to exchange engineering and scientific data becomes increasingly critical. The resurgence of manned exploration efforts to the Moon and Mars further elevates the importance of communications to one of guaranteeing the safety of humanity's explorers. The NASA Deep Space Network, managed and operated by the Jet Propulsion Laboratory (JPL), is an international network of three facilities strategically located around the world to support constant observation of spacecraft launched as part of various interplanetary (and indeed, interstellar) missions. As one of the largest and the most sensitive telecommunications systems in the world, DSN also supports Earth-orbiting missions along with radio astronomy, radar astronomy, and related solar system observations. 

With 12 operational antennas (as of 2019) spread across three locations --- Goldstone, USA, Madrid, Spain and Canberra, Australia --- DSN has served roughly 150 missions for spacecraft communications, and at the time of writing is very near its full capacity. For some weeks, the system is already oversubscribed by the various missions, especially those that cluster in the same portion of the sky \cite{Johnston2008DeepUncertainty}. In addition, the recently-launched Mars 2020 mission adds additional requirements with eight cameras and sophisticated instruments to search for biological evidence of life on the surface. The combined factors of more frequent missions and increased demand for higher-fidelity data operations are expected to significantly increase the load on the DSN. To address the issues of over-subscription, budget constraints and system downtimes, there is urgent need (and thus much ongoing research) to improve the DSN scheduling process such that ``better'' candidate schedules (e.g., with more tracks placed, less conflicts, fairer distribution of tracking time across missions, etc.) can be generated in a much shorter turnaround time. This implies a need to search the solution space for good candidate schedules with such expediency and completeness that exceeds human capabilities, as well as a need to alleviate the bottleneck imposed by the peer-to-peer negotiations process.

\begin{figure}[h]
    \centering
    \includegraphics[width=\linewidth]{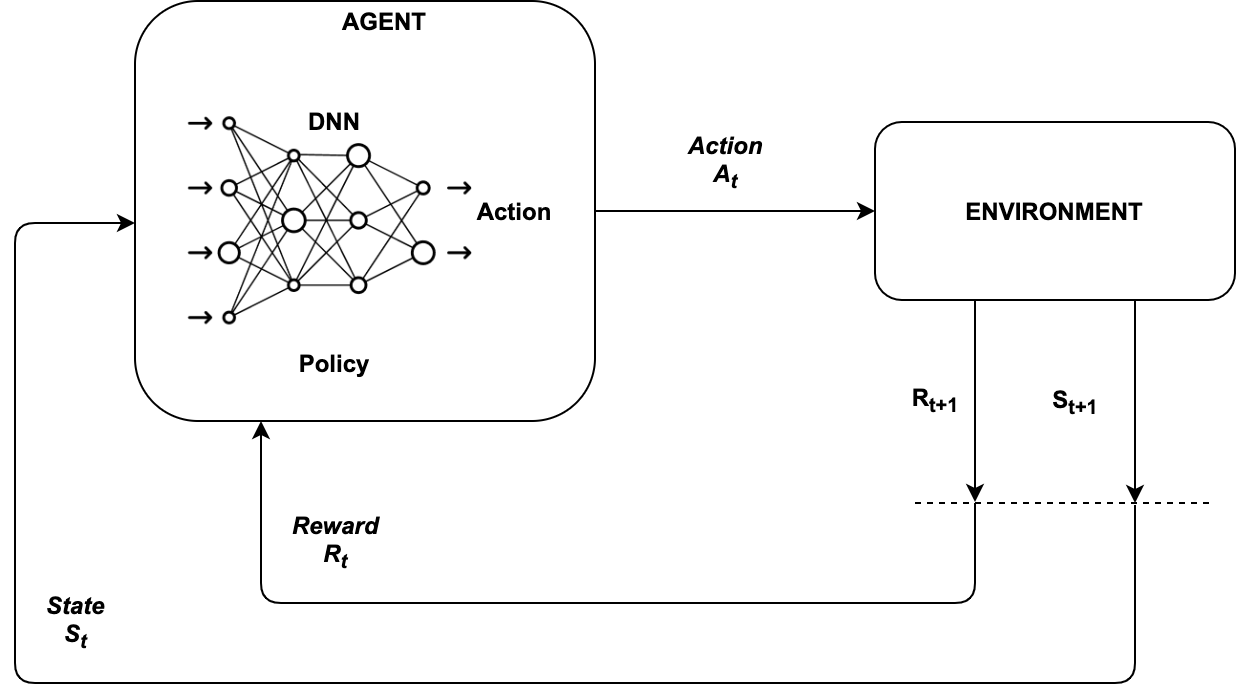}
    \caption{Deep RL Canonical Diagram}
    \label{fig:deeprl}
\end{figure}

Real-world optimization tasks typically have a large number of operational or physical constraints. When solving such problems with conventional operations research techniques, great care is taken to formulate the problem so as to avoid the ``curse of dimensionality'' in which the problem becomes exponentially complex and computationally intractable. The DSN scheduling problem indeed imposes numerous resource allocation constraints due to the wide range of spacecraft orbits, mission requirements, and operational considerations (e.g., hand-off between DSN complexes as the Earth rotates relative to the spacecraft). Deep reinforcement learning (deep RL) is a recent alternative to these conventional approaches that has shown promise in solving complex tasks that are typically considered to rely heavily upon intuition or creativity \cite{DBLP:journals/corr/MnihKSGAWR13,alphago}.
A sub-domain in the field of Artificial Intelligence (AI), deep RL is a combination of reinforcement learning \cite{sutton2018reinforcement} and deep learning. Deep RL is fundamentally represented as a Markov Decision Process (MDP) and typically consists of an agent that interacts with an environment by observing rewards for actions that it takes, as shown in Fig. \ref{fig:deeprl}. 

Recent work on the application of deep RL to scheduling problems in cloud computing resources \cite{Wang2019_MultiAgent} and wireless networks \cite{Wang2019_Cellular} has demonstrated the capability of these algorithms to learn complex rules and strategies required to accomplish such tasks. It has been shown to perform comparably if not better than conventional metaheuristic optimization and search methods on classical operations research problems \cite{Balaji2019}. There have also been multiple instances where deep RL was successfully applied to NASA use cases \cite{zhang2000_jobshop, Rubinsztejn2020, Ferreira2017Multi-objectiveCommunications}. With regards to JPL, the appealing aspects of this approach are as follows:
\begin{itemize}
    \item Upfront investment of training an agent, however high in terms of initial resource requirements, is amortized over future problem sets (weeks) with near real-time inference, which can be performed using consumer hardware. This precludes running classical optimization solvers or training agents from scratch for every scheduling cycle. 
    \item Potential infusion into various other ongoing areas of research such as job shop scheduling \cite{zhang2000_jobshop} and similar use cases at JPL, as well as applications in the public domain. 
\end{itemize}

\subsection{Contributions}
In this paper, we propose a policy optimization based scheduling approach to effectively generate de-conflicted candidate schedules for a given week using mission requests, antenna availability and other constraints as inputs. The purpose of this solution is threefold:
\begin{itemize}
    \item Reduce scheduling turnaround time from a few months to few days.
    \item Increase antenna utilization and thus accommodate more missions.
    \item Minimize the unsatisfied time fraction experienced by each user, i.e., improve ``fairness'' of track allocations across missions.
\end{itemize}

\section{Related Work}
\label{sec:related_work}
Deep Space Network schedules are typically generated a year into the future with allocations to the minute, and are performed manually, one week at a time \cite{Clement2005}. Requested tracks are 1 to 8 hours long and are to be allocated in a view period (VP), defined as the period of time in which the spacecraft is visible to one or more antennas. 
In addition to the set of legal view periods for a given mission, some of the major constraints in DSN scheduling include \textit{quantization} (whether scheduled activities are to occur on 1-minute or 5-minute constraints), sufficient \textit{separation} of contacts (so that onboard data capacity is not exceeded), duration \textit{flexibility} (reduction or extension of tracking time) and \textit{splitting} of requests into multiple tracks \cite{Johnston2014}. Ongoing work seeks to further incorporate the notion of user preferences and mission priorities into the scheduling algorithm such that lower preference requests can be omitted under oversubscription, thereby reducing the amount of peer-to-peer negotiation when potentially high-prority tracks are omitted instead \cite{Johnston2019_UserPref}.

The complexity of the DSN scheduling problem is well-known to the DSN user community, and a large body of literature exists around its solution. Guillaume \textit{et al.} \cite{Guillaume2007} explored a formulation of the problem in terms of evolutionary techniques, and leveraged that formulation to generate a population of Pareto-optimal schedules under varying conflict conditions. More recently, Oller \cite{RuedaOller2019} and Alimo \textit{et al.} \cite{alimo2021} formulated the task as Mixed Integer Linear Programming (MILP) problems to develop scheduling systems (for the long-range and mid-range scheduling problems, respectively) that incorporate many of the DSN's operational and physical constraints. Hackett \textit{et al.} \cite{Hackett2019_DemandAccessScheduling} investigated a beacon-tone demand access scheduling approach, whereby spacecraft, rovers and landers themselves submit ad-hoc requests for tracking time, which are then scheduled in real-time. The authors found that the paradigm decreased the number of required tracks compared to the conventional ``pre-allocated'' approach. On the other hand, \cite{Hackett2019_Thesis} propose multi-objective reinforcement learning cognitive engine using deep neural networks to provide orbit planning and optimization designers the capability to leverage this framework and request resources on-demand. The authors in their other work,  talk about ``demand access'' wherein spacecraft, or rovers request track time on the network themselves using a beacon-tone system and obtain ``on-the-fly'' track time on shared-user block tracks.

\section{Problem Formulation and Design}
\subsection{Input Datasets}
The main dataset used in this work is a set of User Loading Profiles (ULPs) for Week 44 of 2016 (an oversubscribed week), which provides the following information for a given mission:

\begin{enumerate}
    \item The number of tracks requested for that week
    \item The set of requested antenna combinations for these tracks
    \item The requested duration for these tracks
    \item The minimum valid duration for each track (used for splitting tracks into multiple periods)
\end{enumerate}

In order to assign requested tracks to a particular antenna combination during a given week, one needs a set of view periods during which the spacecraft is visible by the requested antenna(s). We use ephemeris data downloaded from JPL's Service Preparation Subsystem (SPS) to assemble, for a given spacecraft and the requested antennas, this set of view periods. This task is a challenge in and of itself because of the potential for multiple-antenna requests that require tracks to be placed on antenna arrays. Such requests necessitate, in addition to the need to identify view periods that overlap across all requested antennas, the need for practical constraints to be taken into account, e.g., minimum duration for the requests, additional setup and teardown times, etc. 

Finally, scheduled maintenance is also taken into account to further constrain the problem. Maintenance data for each antenna is downloaded from SPS and used in the view period identification step to filter out view periods that overlap with maintenance periods for a given antenna. 

The aforementioned input datasets and the overall steps taken to obtain a final problem set to be used in the formulation is shown in Fig. \ref{fig:data_pipeline} below.

\begin{figure}[H]
    \centering
    \includegraphics[width=0.9\columnwidth]{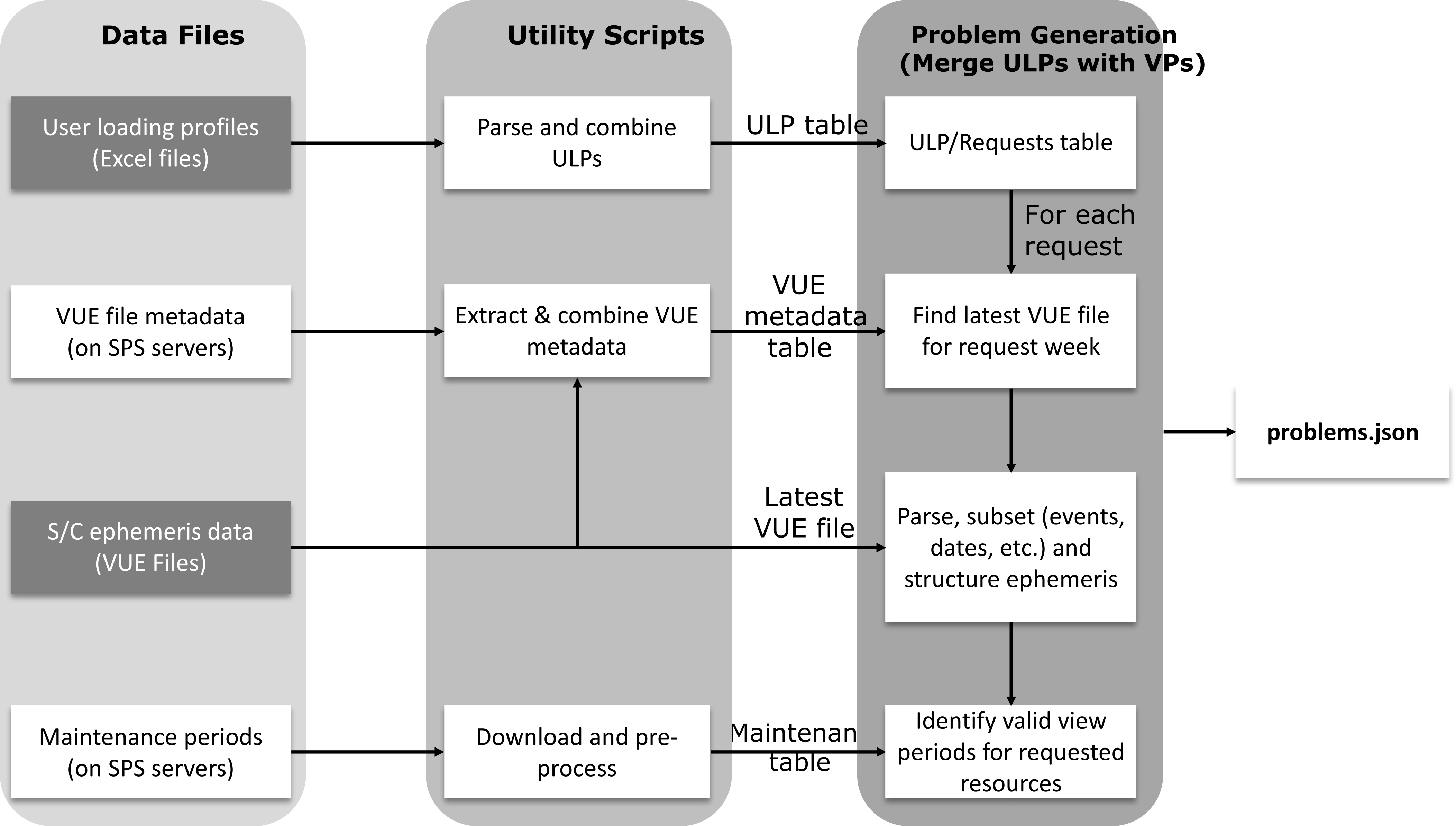}    
    \caption[]{Flow chart illustrating main steps used to generate the problem set, Week 44, 2016 used in this paper.} 
    \label{fig:data_pipeline}
\end{figure}

\subsection{Model/Environment}
This section provides details about the environment used to simulate/represent the DSN Scheduling problem. The environment is implemented according to the \textit{OpenAI Gym} \cite{gym} API in order to maintain compatibility with widely used reinforcement learning libraries such as \textit{RLlib} and \textit{stable-baselines}.

The simulation is instantiated with the problem set generated using the pipeline shown in Fig. \ref{fig:data_pipeline}, as well as a dictionary of DSN antennas. Therefore, episodes in this simulation are centered around week problems. Such a formulation is well-aligned with the DSN scheduling process described in Sec. \ref{sec:related_work}, which generates schedules on a per-week basis.

Each \verb Antenna  object, initialized with start and end bounds for a given week, maintains a list of tracks placed as well as a list of time periods (represented as tuples) that are still available. 
Algorithm \ref{alg:dsn_scheduling_sim} details the general algorithm used in this environment to satisfy requests in the problem set.

\begin{algorithm}[h]
\SetAlgoLined
    \KwData{week problem set (see Fig. \ref{fig:data_pipeline})}
    \While{$n_{rem} > 0$ or $n_{steps} < 2 n_{requests}$}{
        choose a request to allocate\;
        \For{antenna in requested antenna combinations}{
            find and keep only valid VPs\;
        } 
        allocate track on antenna with longest valid VP\;
        \If{duration of VP $>$ requested duration}{ 
            randomly shorten VP to match requested duration\;
        }
        calculate seconds allocated\;
        return reward and observation\;
    }
    \caption{DSN Scheduling Simulation}
    \label{alg:dsn_scheduling_sim}
\end{algorithm}


As seen in the simulation steps detailed in Algorithm \ref{alg:dsn_scheduling_sim}, \verb Antenna  objects provide the capability to process the set of valid view periods identified in Fig. \ref{fig:data_pipeline} according to the antenna's availability and output a set of view periods that do not overlap with existing tracks already placed on that antenna. For multi-antenna requests, these available view periods for each antenna in the array are then passed through an overlap checker to find the overlapping ranges.  

For the view periods that are available, the antenna provides utilities to check whether a view period is valid based on DSN-specific heuristics and rules. For the present work, a view period $\left(t_1, t_2\right)$ with an associated setup/calibration duration $d_s$ and teardown duration $d_t$ is considered valid if all the following conditions return true:
\begin{enumerate}
    \item $\left(t_1 - d_s, t_1\right)$ is available\footnote{i.e., does not overlap any of the tracks already placed on this antenna}, or if $(t_2 - t_1) \geq d_{min} + d_s + d_t$ 
    \item $\left(t_2, t_2 + d_t\right)$ is available, or if $(t_2 - t_1) \geq d_{min} + d_s + d_t$ 
    \item $(t_2 - t_1) \geq d_{min}$, where $d_{min}$ is the minimum requested duration for this track
\end{enumerate}


As we will discuss in the following sections, the present environment handles most of the ``heavy-lifting'' involved in actually placing tracks on a valid antenna, leaving the agent with only one responsibility --- to choose the ``best'' request at any given time step.  The simulation described thus far is a preliminary implementation. Constraints such as the splitting of a single request into tracks on multiple days or Multiple Spacecraft Per Antenna (MSPA) are important aspects of the DSN scheduling problem that require experience-guided human intuition and insight to fulfill. Being cognizant of this limitation, we intentionally implement this environment in a modular fashion such that subclasses with additional constraints can be easily defined in the future.

\subsection{State Space/Observation}

At any given point in the simulation, the environment keeps track of:

\begin{itemize}
    \item the distribution of remaining requested durations, 
    \item the total outstanding requested hours for that week,  
    \item the number of unique missions with outstanding requests,  
    \item the remaining number of requested tracks, and  
    \item the number of remaining free hours on each antenna.  
\end{itemize}  

In order to use the same observation space over multiple weeks, we specify a bound on the maximum number of requests (i.e., requested tracks) that are valid in any given week. For requests in the year 2016, a bound of 500 provided sufficient margin across all weeks. Thus 500 entries are defined for the distribution of remaining requested durations. 

This state space of the environment is represented as a 1-D array that indicates the number of remaining unique missions, the number of remaining requests, the total remaining duration requested, as well as the remaining unallocated duration in each request. 

\subsection{Action Space}
There are multiple ways to enumerate the actions a reinforcement learning agent can take at each time step. An initial attempt specified the action space as a 2D binary grid whose rows represented the individual DSN antennas and the columns represented discretized time periods. When flattened/reshaped into a 1-D array, this resulted in a formidable action space of size $2^{M \times K}$ where $M$ is the number of DSN antennas and $K$ is the number of time steps resulting from the discretization of the entire week by a given time step. Since such a large action space precludes efficient learning and makes the addition of DSN-defined constraints difficult, the current iteration of the action space for the DSN scheduling environment is intentionally simple --- a single integer that defines which item in a request set the environment should allocate. Action masking is used in order to prevent the agent from choosing requests that have already been satisfied.

This implementation was developed with future enhancements in mind, eventually adding more responsibility to the agent such as choosing the resource combination to use for a particular request, and ultimately the specific time periods in which to schedule a given request. These decisions are hierarchical in nature and resemble the possible actions for each Dota agent in OpenAI Five \cite{openai_five}, whereby an agent would for instance decide to attack, select a target to attack, and decide whether to offset the action in anticipation of the target unit's future position.

\subsection{Rewards}
In the DSN scheduling environment, an agent is rewarded for an action if the chosen request index resulted in a track being scheduled. Here, the reward is given by,
\begin{equation}
    r_t(s, a) = \frac{T_{allocated}}{T_{requested}}
\label{eq:reward_func}
\end{equation}
where $T_{allocated}$ is the total time scheduled across all antennas for this request and $T_{requested}$ is the requested time allocation for the entire week. 

At each time step, the reward signal is a scalar ranging from 0 (if the selected request index did not result in the allocation of any new tracking time) to 1 (if the environment was able to allocate the entire requested duration). As one can surmise, the theoretical maximum reward that can be achieved in an episode is the \textit{number of requests in that week}. 

\subsection{Training Algorithm}

For this preliminary exploration, we use the Proximal Policy Optimization (PPO) algorithm \cite{Schulman2017_PPO} implemented in the RLlib reinforcement learning library \cite{Liang2018_rllib}. While Schulman \textit{et al.} demonstrated state-of-the-art performance with PPO on robotic locomotion/optimal control and Atari game playing, the algorithm has been shown to be feasible on stochastic optimization problems in operations research \cite{Balaji2019}. Furthermore, an RL agent trained on REINFORCE --- another policy gradient algorithm similar to PPO --- was shown to perform similarly and sometimes better than existing heuristics-based approaches for scheduling multi-resource clusters \cite{Mao2016ResourceLearning}. 

RLlib implements PPO in an actor-critic fashion. The actor is a typical policy network that maps states to actions, whereas the critic is a value network that predicts the state's value, i.e., the expected return for following a given trajectory starting from that state. 
For a batch of observations from the environment, the actor network predicts a distribution over the set of available actions. The training algorithm then samples a specific action from this distribution based on a given exploration strategy.

After an action is selected, the critic estimates the \textit{advantage} $A_t(s, a)$ as a function of the (temporal-difference) error $\delta_t$ between the value function predicted by the network and the actual rewards returned by the environment. The error term is defined as
\begin{equation}
    \delta_t = r_t + \gamma V(s_{t+1}) - V(s_t)
\end{equation}
where $V$ is the critic's current model of the value function, and $r_t$ is the ratio of action probabilities for the current state $s_t$ under the current policy to the action probabilities for $s_t$ under the old policy.

Thus for a given policy defined by the parameters $\theta$, the objective used in PPO is as follows,
\begin{equation}
    L^{PPO}(\theta) = \hat{\mathbf{E}}\left[\min\Big(r_t(\theta)A_t,\  \text{clip}(r_t(\theta), 1 - \epsilon, 1 + \epsilon)A_t\Big)\right]
\label{eq:ppo}
\end{equation}
where $\epsilon$ is a hyperparameter proposed in \cite{Schulman2017_PPO} to clip $r_t$ and thus prevent large policy updates that result in irrecoverable decreases in agent performance. Since the gradient of Eq. \ref{eq:ppo} is an estimator for the policy gradient, using this loss function as the objective to a stochastic gradient ascent problem is a surrogate for updating the policy to encourage good actions and weaken the tendency for actions that perform worse than expected.


While the results in \cite{Schulman2017_PPO} are obtained using an actor and critic that share the same layers, the neural architecture used in this work is one that has separate layers (and thus parameters) for both the policy and the value function. Throughout all experiments, we use a fully-connected neural network architecture with 2 hidden layers of 256 neurons each. Based on the observation/state space defined above, the input layer is of size 518; the first three entries are the remaining number of hours, missions, and requests, the following set of 500 entries are the remaining number of hours to be scheduled for each request, and the final 15 entries are the remaining free hours on each antenna. We use a maximum number of requests of 500 to ensure that the same observation space can be used across multiple weeks. 

\begin{figure}[H]
    \centering
    \includegraphics[width=0.9\columnwidth]{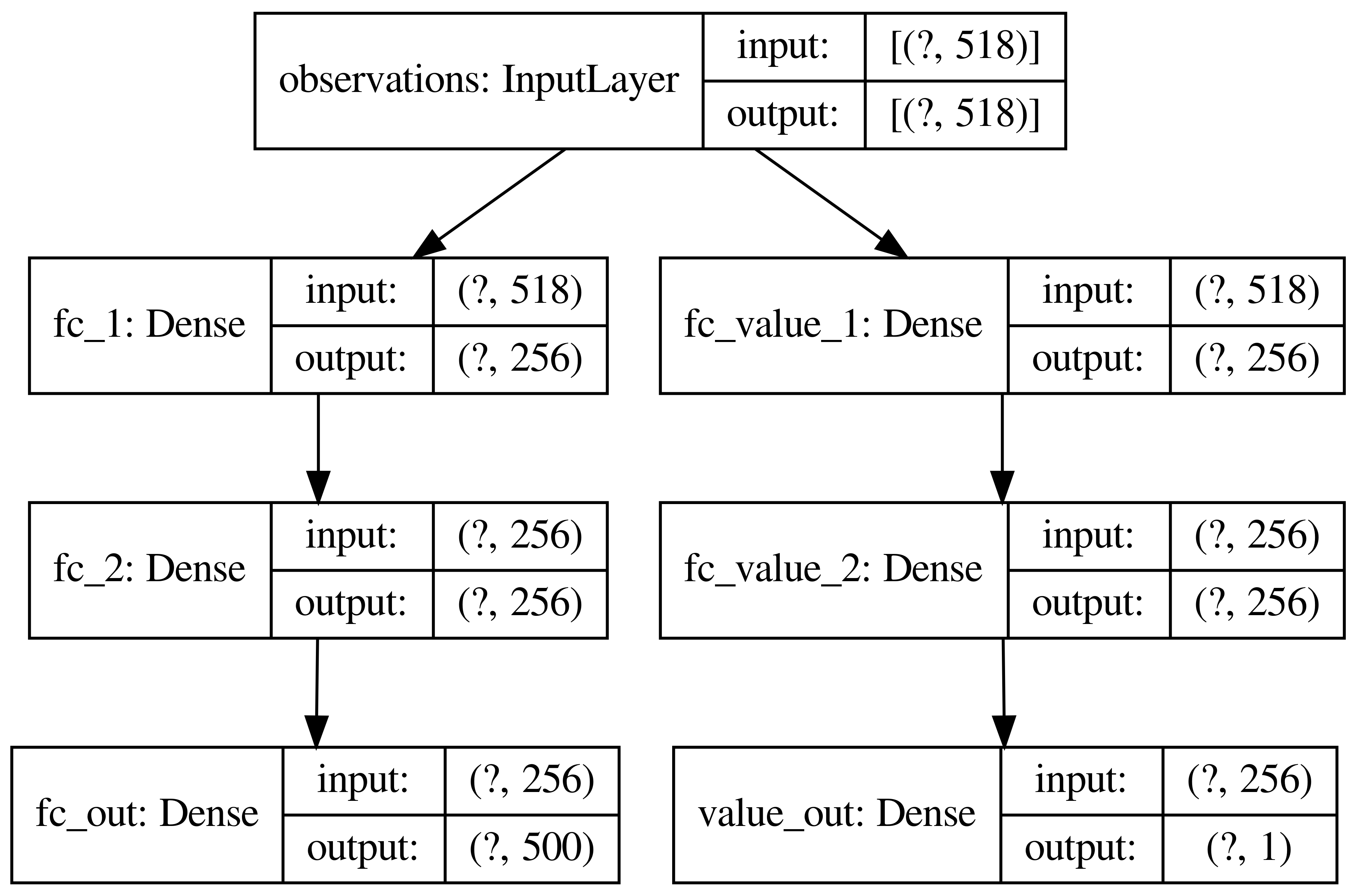}
    \caption[]{Actor-critic network architecture used in this work. The left branch represents the actor network which maps observations to actions, whereas the right branch depicts the actor which learns to estimate the value of a given state.}
    \label{fig:model}
\end{figure}

\section{Results and Discussion}

\begin{figure*}[!bt]
    \centering
    \begin{subfigure}{0.3\textwidth}
        \includegraphics[width=1.0\linewidth]{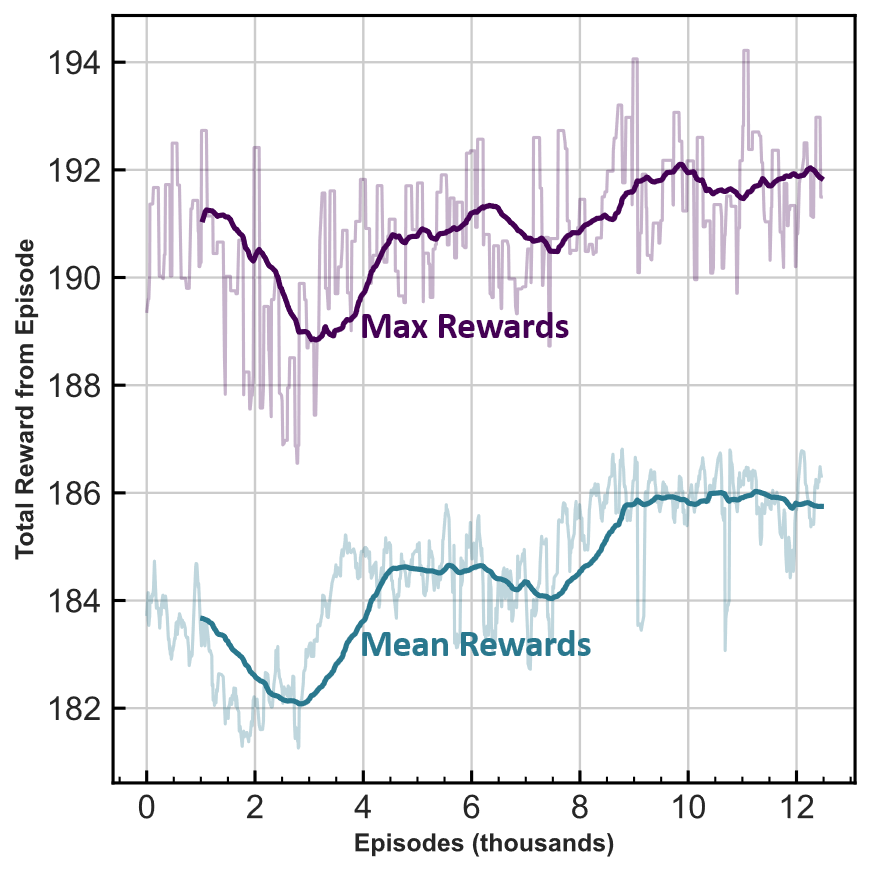}
        \caption[]{Mean and maximum rewards}
        \label{fig:training_rewards}
    \end{subfigure}
    \begin{subfigure}{0.3\textwidth}
        \includegraphics[width=1.0\linewidth]{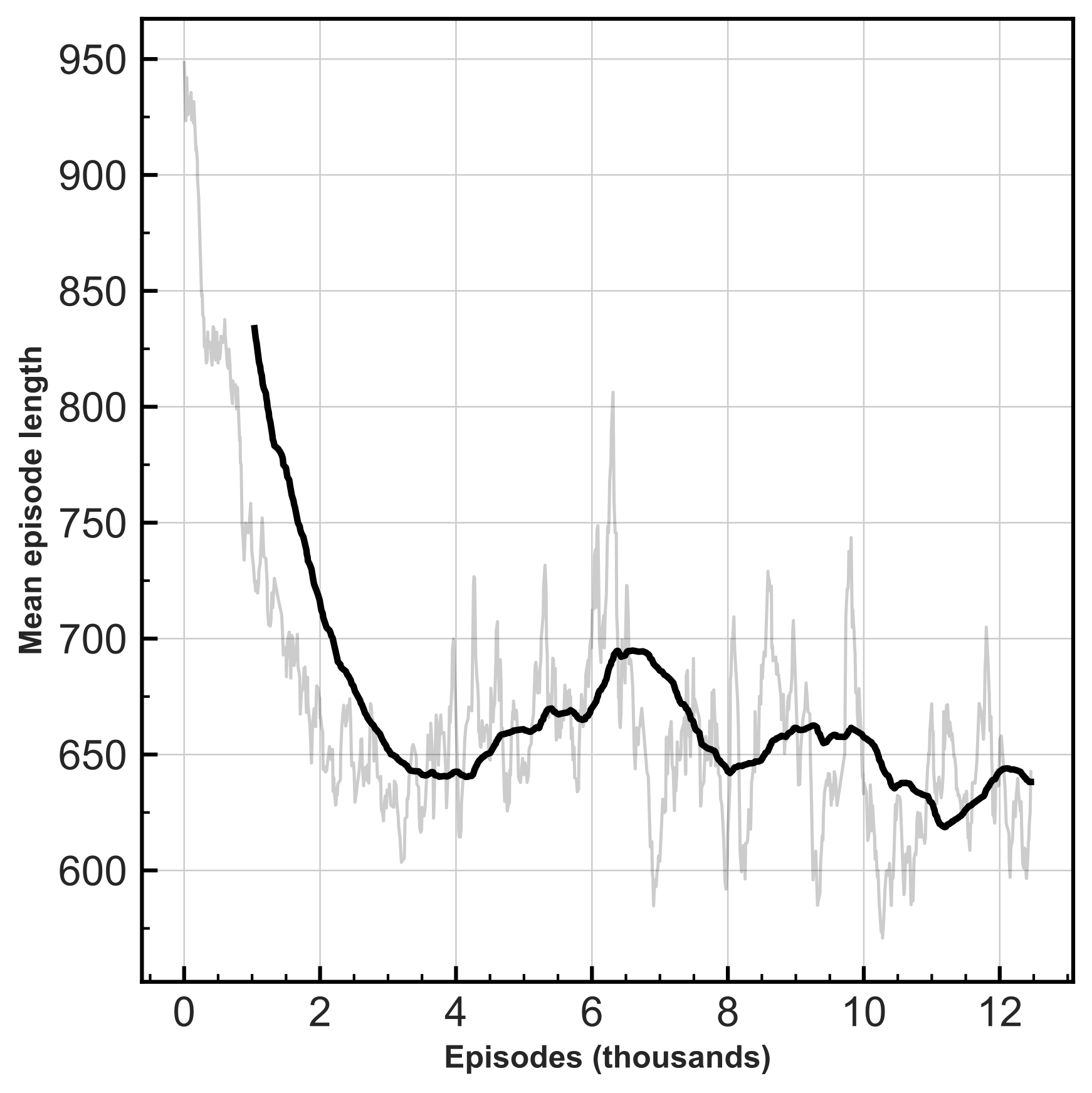}
        \caption[]{Average number of steps taken}
        \label{fig:training_episode_length}
    \end{subfigure}
    \begin{subfigure}{0.3\textwidth}
        \includegraphics[width=1.0\linewidth]{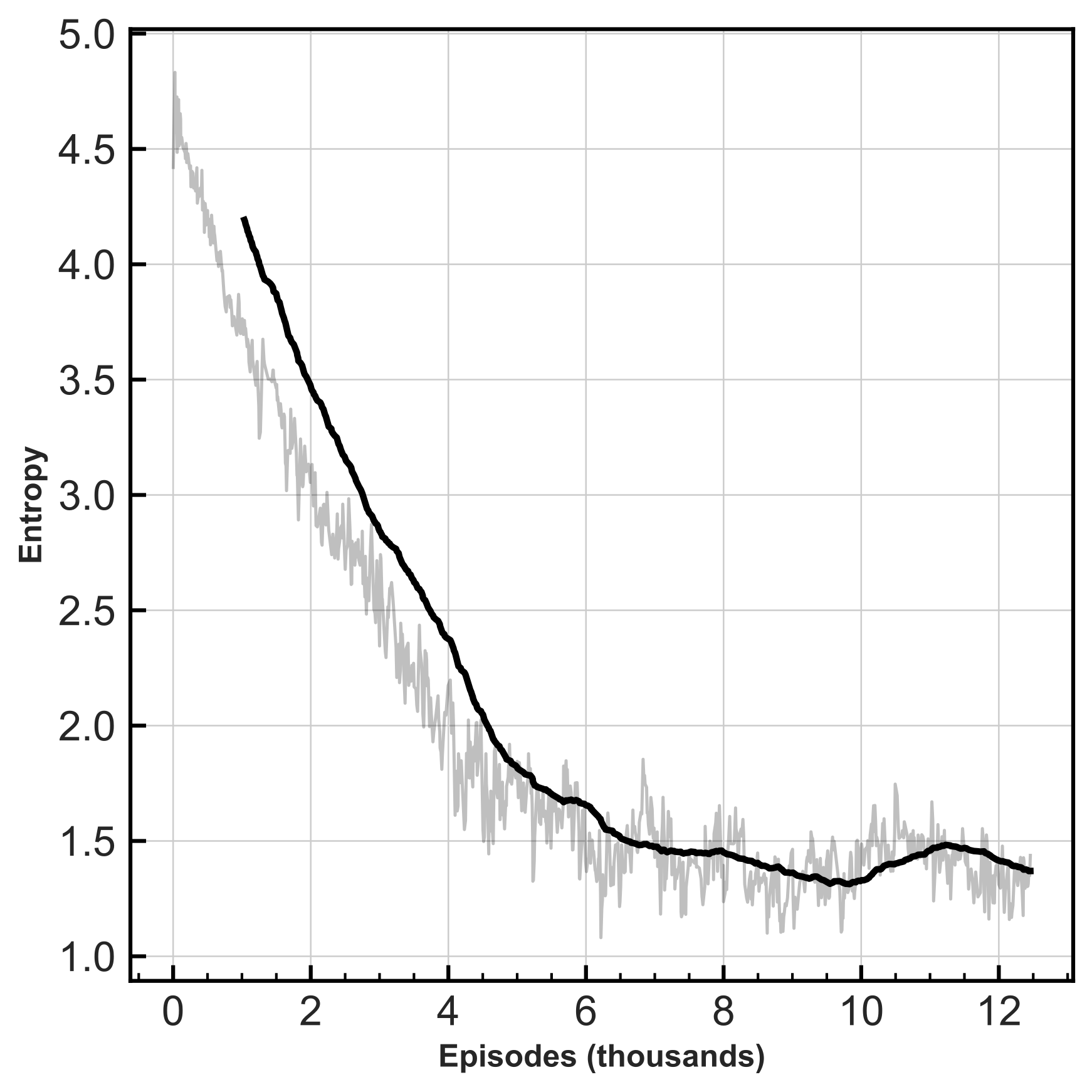}
        \caption[]{Entropy}
        \label{fig:training_entropy}
    \end{subfigure}
    \caption{Evolution of key metrics during PPO training of the DSN scheduling agent. Rewards and episode length statistics were calculated across 20 evaluation episodes.}
    \label{fig:training_metrics}
\end{figure*}

In this section, we first present details of the training process as well as the hyperparameters used. We then present preliminary solutions obtained using the formulation described above and compare those solutions with that of an agent taking random solutions. Solutions are presented for Week 44 of 2016. Excluding maintenance requests on the individual antennas, the DSN received a total of 286 requests for that week, which amounted to 1,770 hours to be allocated across DSN's antennas. 

\subsection{Experimental Setup}

Training was performed on a single Amazon EC2 instance with 4 GPUs and 32 CPUs, and the agent was trained for roughly ${\sim}$10M time steps using the \textit{RLlib} framework. \textit{RLlib} provides trainer and worker processes --- the trainer is responsible for policy optimization by performing gradient ascent while workers run simulations on copies of the environment to collect experiences that are then returned to the trainer. \textit{RLlib} is built on the \textit{Ray} backend, which handles scaling and allocation of available resources to each worker. 

PPO uses Stochastic Gradient Descent (SGD) algorithm, and in this experiment we set minibatch size to 128 and the number of epochs to 30 for optimizing the surrogate objective given in Eq. \ref{eq:ppo}. While learning rate schedules can be defined in \textit{RLlib}, the results presented here were trained using a constant learning rate of 5\text{e}-5. The target Kullback–Leibler (KL) divergence \cite{kullback1951} is set to 0.01 and the Generalized Advance Estimator (GAE) parameter, $\lambda$, is set to 1.0. $\lambda$ is a bias-variance tradeoff parameter; higher values imply higher variance \cite{schulman2018highdimensional}. The discount factor or gamma parameter is set to 0.99, which gives more weighting on long-term rewards rather than immediate rewards. The clipping parameters for PPO policy and value function loss are set to RLLib defaults, and critic baseline is set to true for making use of GAE. 

Fig. \ref{fig:training_metrics} shows the evolution of several key metrics from the training process. In Fig. \ref{fig:training_rewards}, mean and maximum rewards achieved by the policy across several 20 evaluation episodes are shown to increase in a stepwise fashion as the number of training episodes increases. One would expect the distribution of rewards to shift rightwards as the policy is progressively updated. Decreases in reward indicate periods where the agent doesn't \textit{exploit} the best-available policy at the time, but instead explores other policies\footnote{Recall that policies in this case are deep neural networks parameterized by $\theta$.} 
to prevent itself from being trapped in local extrema.
Furthermore, the average number of steps taken in each episode (Fig. \ref{fig:training_episode_length} is shown to decrease with training, indicating that the agent is capable of achieving better-performing schedules without spending additional steps to select requests that cannot be allocated. In other words, this may be an indication that the agent is learning to prioritize requests that \textit{can} be allocated by the environment based on the availability of the antennas. 
Finally, Fig. \ref{fig:training_entropy} shows the evolution of entropy as training progresses. Entropy is an important indicator of whether there is variance in the actions taken by the policies being trained. The gradually decreasing entropy in Fig. \ref{fig:training_entropy} indicates that the PPO algorithm is converging on an optimal policy while maintaining its exploration policy.

\subsection{Random Agent Baseline}
Due to complexities in the DSN scheduling process described in Section \ref{sec:intro}, the current iteration of the environment has yet to incorporate all necessary constraints and actions to allow for an ``apples-to-apples'' comparison between the present results and the actual schedule for week 44 of 2016. For example, the splitting of a single request into multiple tracks is a common outcome of the discussions that occur between mission planners and DSN schedulers. This allows for tracks to be fit into gaps that full requests otherwise would not, at the cost of increased overhead time due to setup and teardown. 

\begin{figure}[!htb]
    \centering
    \includegraphics[width=0.9\columnwidth]{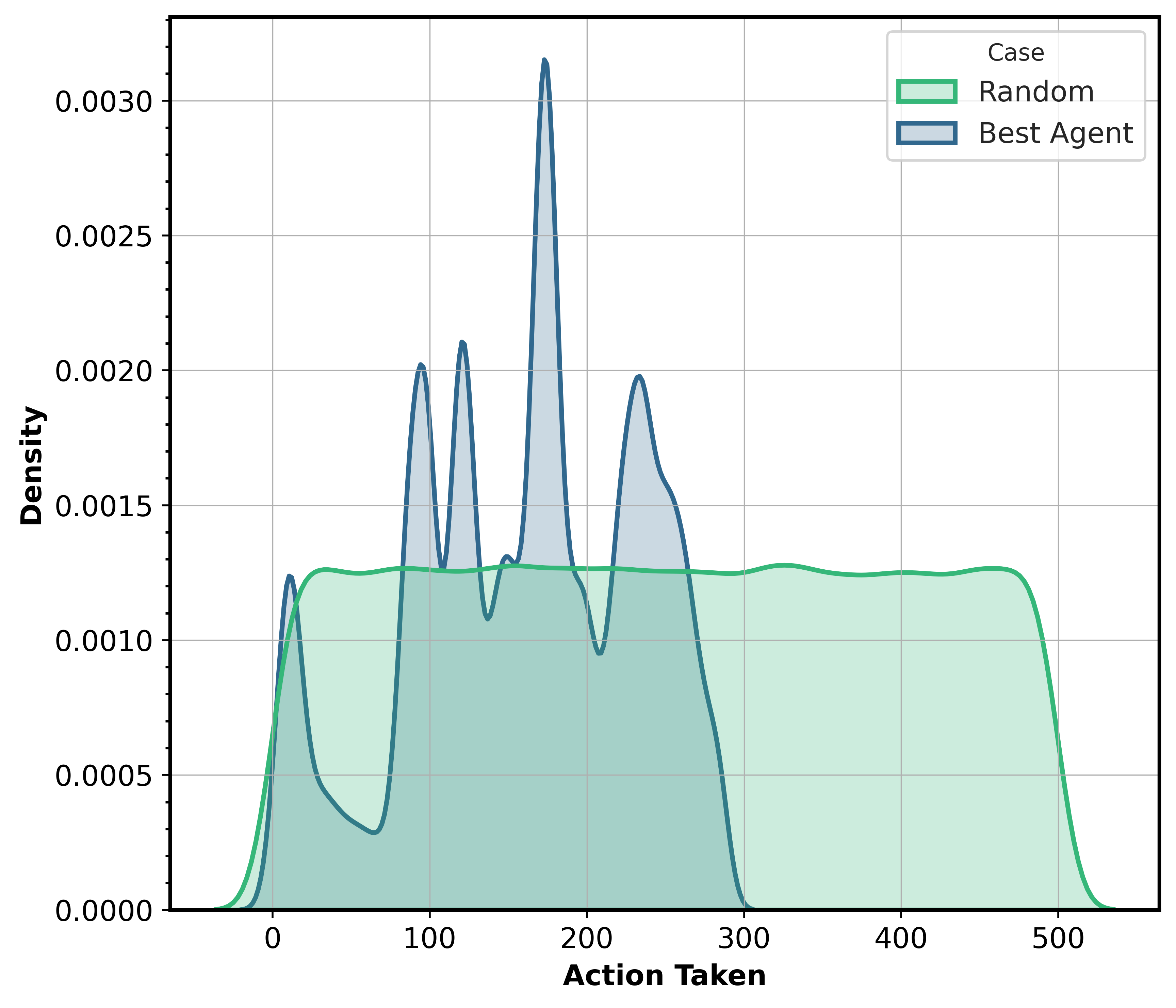}
    \caption[]{Kernel density estimate of actions taken over 100 episodes for the random agent (green) and best agent (blue). Note that episodes consist of multiple steps, and results here are shown for actions selected by the agent at each step.}
    \label{fig:action_dist}
\end{figure}

Instead of comparing to historical data, we define the performance of a random agent\footnote{A random agent is one that uniformly samples the action space at every time step of the environment.} 
to be the baseline result. Recall that actions in this case are integers that represent the index of the request to schedule) and passing them into the environment. As seen in Fig. \ref{fig:action_dist}, a random agent without action masking chooses uniformly across the entire range of possible request indices (0-500). 

\begin{figure}[!htb]
    \centering
    \includegraphics[width=0.9\columnwidth]{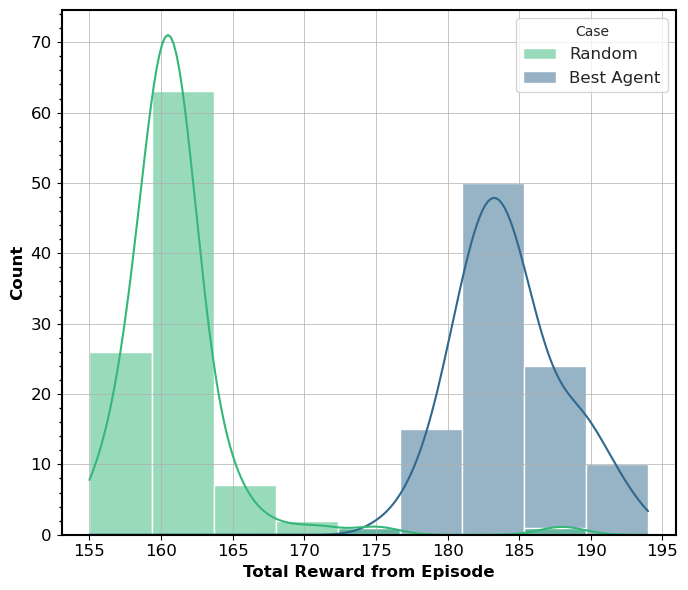}
    \caption[]{Distribution of total rewards obtained over 100 episodes for the random agent (green) and best agent (blue). The reward distribution achieved by the trained agent exhibits an obvious shift to the right, indicating learning by the agent.}
    \label{fig:reward_dist}
\end{figure}

\subsection{Comparison with Trained Agent}

The agent with the best performance (mean rewards in Fig. \ref{fig:training_rewards}) was chosen as our preliminary benchmark against the random baseline. 
This was the agent with the policy that had undergone roughly 700 SGD updates, or roughly 10,000 episodes. 
We perform 100-episode rollouts/evaluations using the best-performing agent and the random agent to sample the stochastic policies.
The action distributions across all episodes shown in Fig. \ref{fig:action_dist} illustrate that action masking indeed keeps agent actions to within the 286 requests for week 44 of 2016. 
Furthermore, Fig. \ref{fig:action_dist} shows a distinct distribution of actions, indicating that, there are requests that the agent ``prefers'' to allocate as opposed to a uniform sampling of the action space.

From the 100 episodes, we extract schedules from the episodes with total rewards closest to the mean reward (${\sim}$161 for the random agent and ${\sim}$184 for the trained agent). Key performance metrics for DSN schedules include the RMS of the \textit{unsatisfied time fraction} across all missions, $U_{RMS}$, maximum unsatisfied time fraction among all missions $U_{max}$ and \textit{antenna utilization}, $A$. These are defined in Eqs. \ref{eq:unsatisfied}--\ref{eq:antenna_utilization}.
\begin{align}
    U_i &= \frac{T_{R_i} - T_{S_i}}{T_{R_i}} \label{eq:unsatisfied}\\
    U_{RMS} &= \sqrt{\frac{1}{N}\sum_{i}^{N}{U_i^2}} \label{eq:u_rms}\\
    U_{max} &= \max_i\left(U_i\right) \label{eq:u_max}
\end{align}
where $T_{R_i}$ represents the total tracking time requested by the $i$-th mission, and $T_{S_i}$ represents the total duration scheduled across all antennas for that mission.

$U_{max}$ is an indication of which mission has the most requests unsatisfied, while $U_{RMS}$ provides a measure of uniformity in allocations over all missions. 

\begin{equation}
    A = \frac{\text{total time antennas not idle}}{\text{total available antenna time for time period}}
    \label{eq:antenna_utilization}
\end{equation}

\begin{figure*}[!htbp]
    \centering
    \begin{subfigure}{0.45\textwidth}
        \includegraphics[width=1.0\linewidth]{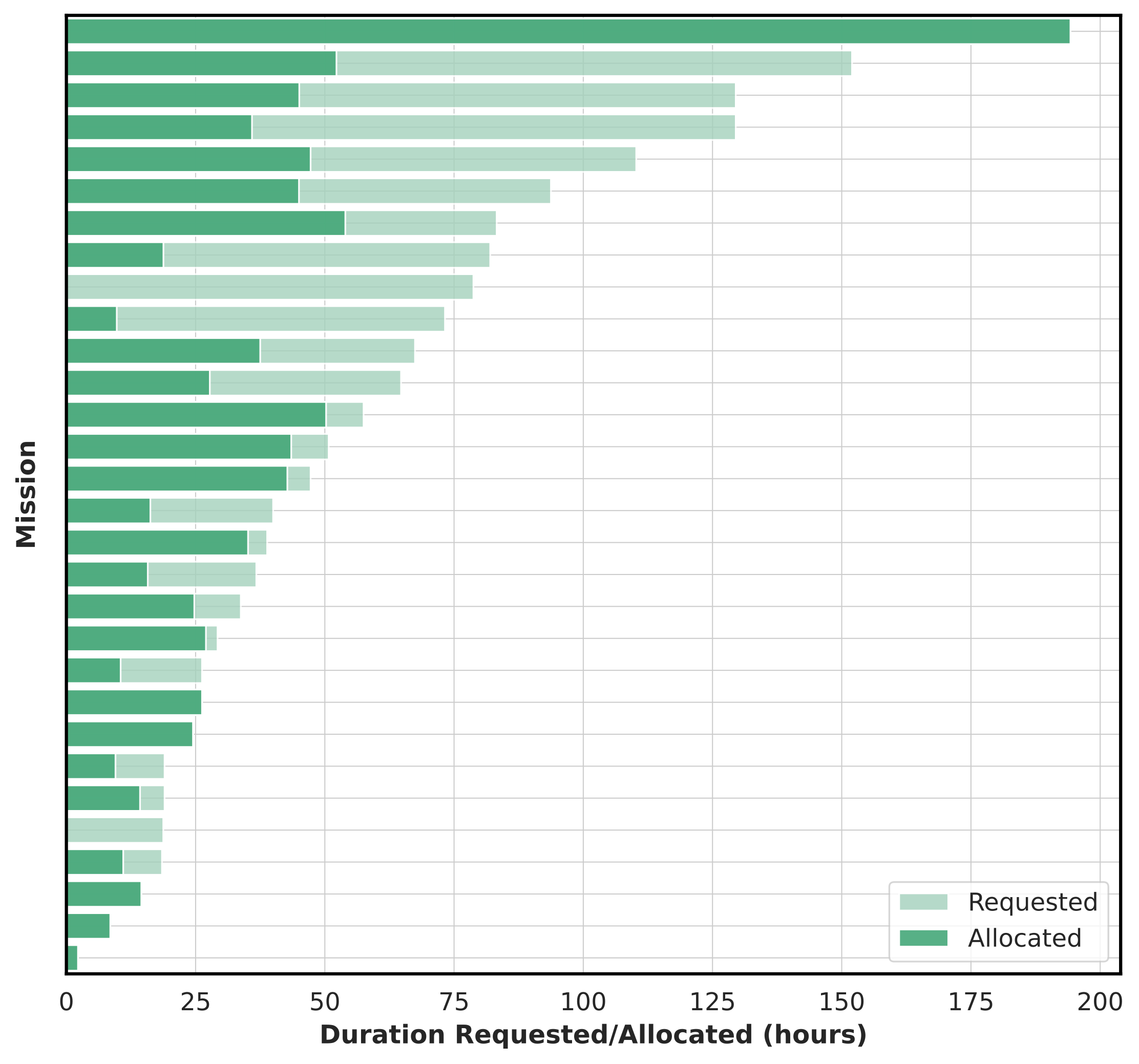}
        
        \caption[]{Random Agent}
        \label{fig:random_num_hours}
    \end{subfigure}
    \begin{subfigure}{0.45\textwidth}
        \includegraphics[width=1.0\linewidth]{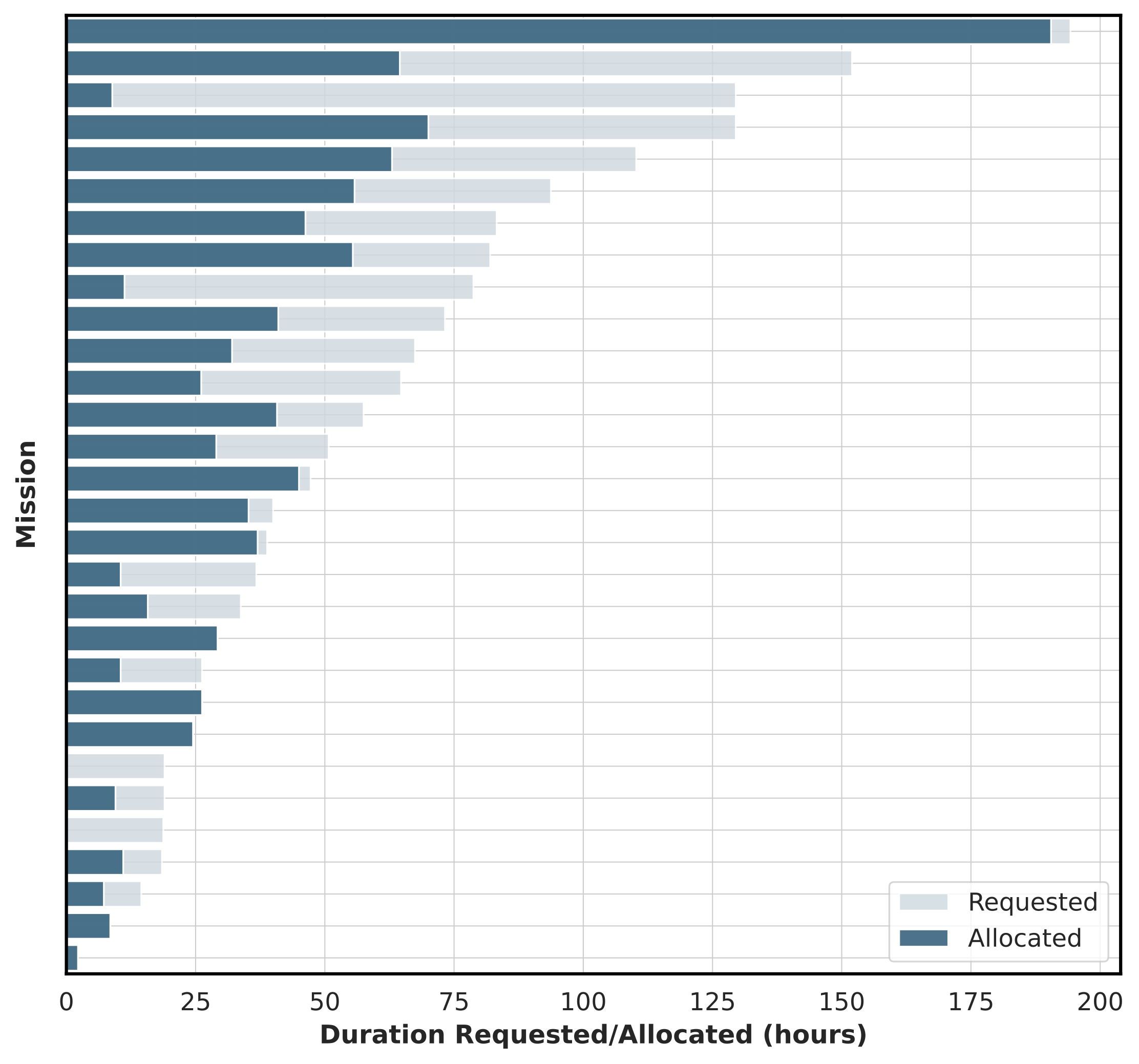}
        
        \caption[]{Trained Agent}
        \label{fig:agent_num_hours}
    \end{subfigure}
    \caption{Comparison of number of hours allocated across all missions using the random and trained agents.}
    \label{fig:num_hours_comparison}
\end{figure*}

\begin{figure*}[!htbp]
    \centering
    \begin{subfigure}{0.45\textwidth}
        \includegraphics[width=1.0\linewidth]{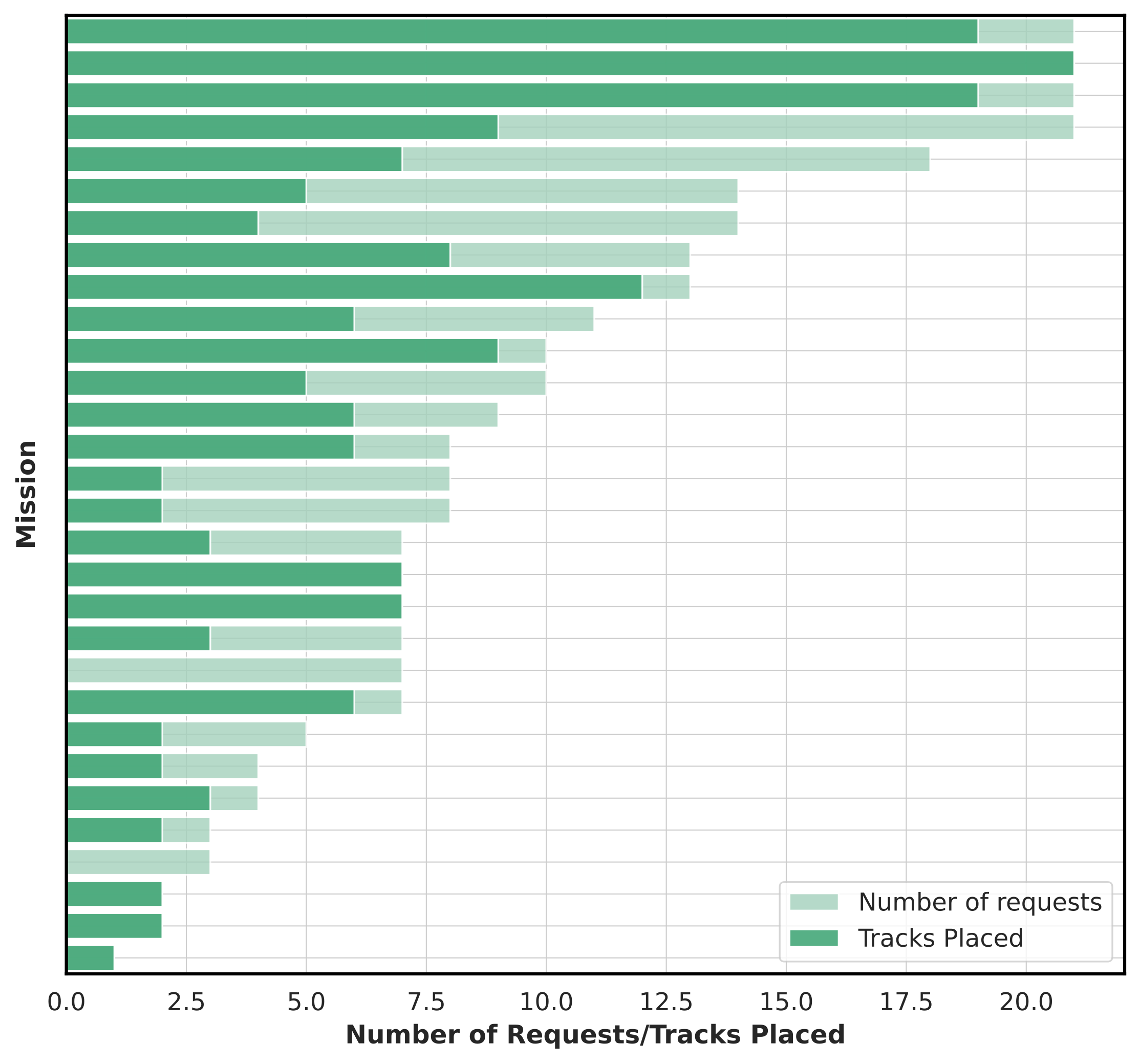}
        
        \caption[]{Random Agent}
        \label{fig:random_num_req}
    \end{subfigure}
    \begin{subfigure}{0.45\textwidth}
        \includegraphics[width=1.0\linewidth]{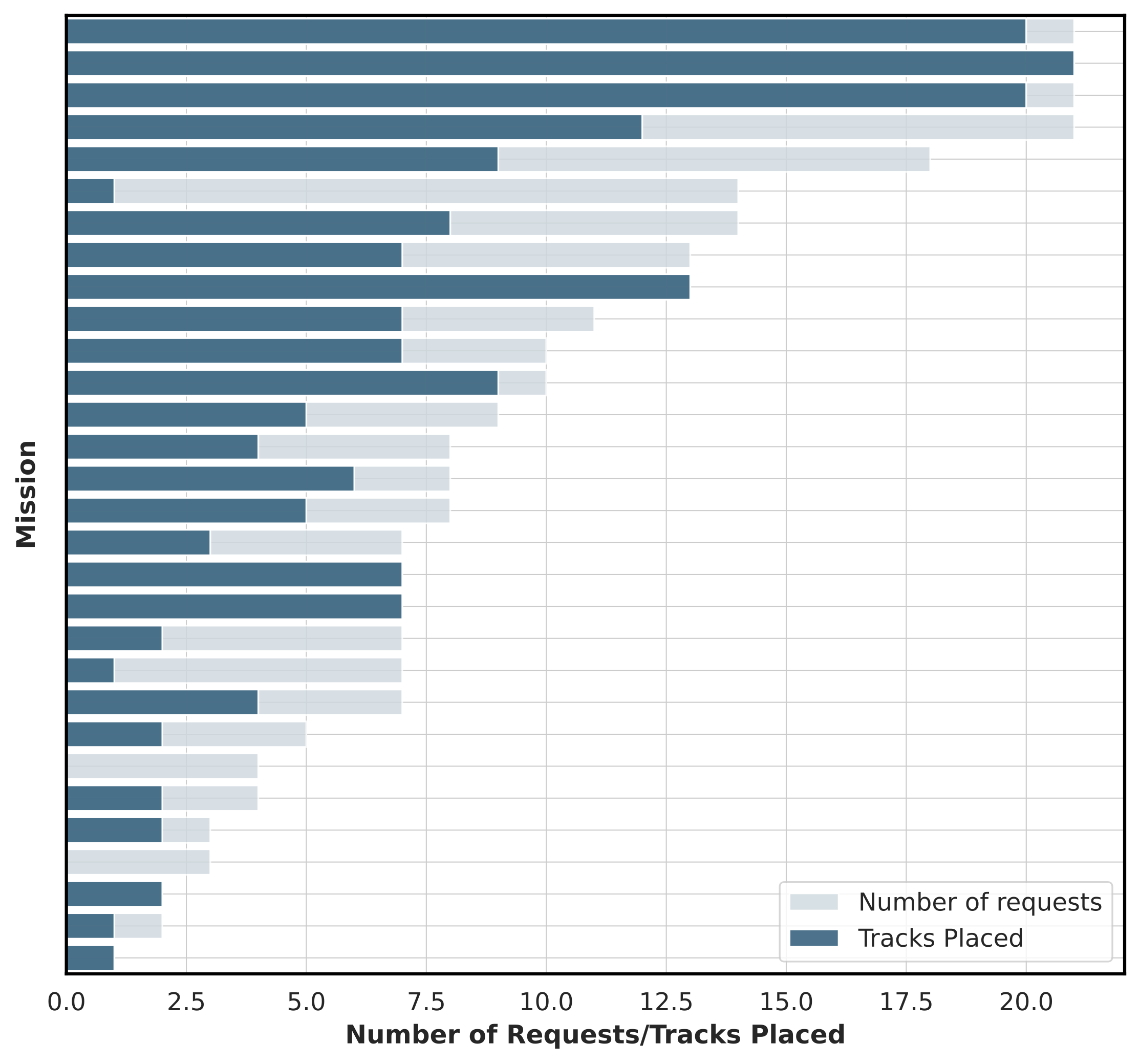}
        
        \caption[]{Trained Agent}
        \label{fig:agent_num_req}
    \end{subfigure}
    \caption{Comparison of number of requests scheduled across all missions using the random and trained agents.}
    \label{fig:num_req_comparison}
\end{figure*}

As seen in Table \ref{tab:schedule_comparison}, the trained agent manages to satisfy 1,007 hours out of the requested 1,770 hours whereas the random agent satisfies 944 hours. Likewise, the trained agent allocates slightly more requests than the random case. The difference in $U_{RMS}$ between the two cases is negligible. Figs. \ref{fig:num_hours_comparison} and \ref{fig:num_req_comparison} show a comparison across 30 missions for the number of hours and number of tracks requested/allocated, respectively. The mission names have been omitted from these figures. 

\begin{table}[H]
\centering
\caption{Comparison of scheduled results using the mean performance of the random agent and the mean performance of the trained agent for Week 44, 2016. }
\begin{tabular}{@{}lcc@{}}
\toprule
\textbf{Agent (Mean performance from Fig. \ref{fig:reward_dist})} & \textbf{Random} & \textbf{Trained} \\ \midrule
Hours satisfied                                & 944             & 1007             \\
Mean satisfied time fraction (\%)              & 60.5            & 59.4             \\
Number of satisfied requests                   & 180             & 188              \\
Mean satisfied request fraction (\%)           & 62.9            & 65.7             \\
RMS of unsatisfied time fraction, $U_{RMS}$ (\%)          & 4.3             & 3.9              \\ \bottomrule
\end{tabular}
\label{tab:schedule_comparison}
\end{table}

The results presented above indicate that, while the agent is definitely learning to choose specific requests to have the environment allocate, the final output schedules exhibit only a modest improvement from randomly chosen actions. This is not surprising considering the simplicity of the agent's action space and the greedy fashion in which the environment allocates requests after receiving an index from the agent. In addition to demonstrating the feasibility of deep RL for scheduling spacecraft communications, the main accomplishment in this work is the implementation of a simple yet modular representation of the DSN scheduling problem within the deep RL framework that can be augmented with increasingly more realistic constraints and more complex RL agents. We discuss promising avenues of research in the next section.


\section{Conclusions and Future Work}
\label{sec:conclusions}
In this paper, we presented a formulation of the DSN scheduling process as a reinforcement learning problem. An environment that encapsulates the dynamics of the scheduling problem was implemented, with the observation space being a series of quantities that represent the state of the remaining problems and the DSN antennas' availability. The agent's action space was simplified for this preliminary task --- a single integer that represents the index to a list of requests for the week. Given this index, the environment then attempts to allocate the request in a greedy fashion, i.e., on the requested antenna combination with the most available time remaining.

Using the aforementioned deep RL formulation with the proximal policy optimization algorithm, an agent was trained on user loading profiles from 2016 for roughly 10M steps. Preliminary results demonstrate observable improvement in agent performance as the underlying policy converges on an optimal policy.
Due to the preliminary nature of this implementation and the complex human-in-the-loop nature of the scheduling process, comparisons could only be performed against a random agent baseline rather than the actual scheduling outcomes. These comparisons indicate that the trained agent exhibits demonstrably more reliable performance than a random agent due to the improved policy, although the absolute gains in schedule-related metrics such as unsatisfied time fraction are small. 

The low performance observed in the trained agent is, perhaps unsurprisingly, due to the simplicity with which the environment and agent were designed. Indeed, it is this intentional simplicity that allows us to leverage the explainability of the agent's progress and learnings rather than performance at this juncture. This cognizance led to very careful planning of the system's implementation such that additional improvements can be made with minimal effort. Thus in ongoing research, we plan to incorporate realistic constraints elicited from requirements discussions while also scaling the datasets to represent complexity of real-world requests. We intend to improve the formulation of action spaces such that the agent also learns to split, shorten and drop tracks wherever necessary, and learn action space representations using action embeddings \cite{Mao2018}. Currently, complexity of input datasets that the agent is being trained on has remained fairly high since we consider the oversubscribed weeks. Though the results demonstrate agent's learning capabilities, neural networks, similar to humans, benefit from gradual increase in the difficulty of the concepts it can learn \cite{Elman1993LearningAD}. To that end, we plan to integrate curriculum learning \cite{Bengio_curriculumlearning} and scale the training examples gradually using curriculum-based training strategies.

%


%






\section*{Acknowledgments}
This effort was supported by JPL, managed by the California Institute of Technology on behalf of NASA. The authors would like to thank JPL Interplanetary Network Directorate and Deep Space Network team, and internal DSN Scheduling Strategic Initiative team members Alex Guillaume, Shahrouz Alimo, Alex Sabol and Sami Sahnoune. U.S. Government sponsorship acknowledged.

\bibliographystyle{IEEEtran}
\bibliography{references.bib}

\end{document}